\documentclass{article}
\usepackage{spconf}
\usepackage{color}
\usepackage{times}
\usepackage{epsfig}
\usepackage{graphicx}
\usepackage{amsmath}
\usepackage{amssymb}

\usepackage{cite}
\usepackage{amsfonts}
\usepackage{algorithmic}
\usepackage{textcomp}
\usepackage{subfig}
\usepackage{wrapfig}

\newcommand{\e}{\mathbf{e}}

\newcommand{\m}{\mathbf{m}}

\newcommand{\x}{\mathbf{x}}

\newcommand{\F}{\mathbf{F}}

\newcommand{\G}{\mathbf{G}}

\newcommand{\Lb}{\mathcal{L}}

\begin{document}

\title{Image Inpainting for High-Resolution Textures using \\CNN Texture Synthesis}

\name{Pascal Laube, 
		Michael Grunwald, 
		Matthias O. Franz \& 
		Georg Umlauf}
\address{Institute for Optical Systems, University of Applied Sciences Constance, Germany\\
 \tt\small plaube@htwg-konstanz.de}

\maketitle

\begin{abstract}
Deep neural networks have been successfully applied to problems such as image segmentation, image super-resolution, coloration and image inpainting.
In this work we propose the use of convolutional neural networks (CNN) for image inpainting of large regions in high-resolution textures.
Due to limited computational resources processing high-resolution images with neural networks is still an open problem.
Existing methods separate inpainting of global structure and the transfer of details, which leads to blurry results and loss of global coherence in the detail transfer step.
Based on advances in texture synthesis using CNNs we propose patch-based image inpainting
by a CNN that is able to optimize for global as well as detail texture statistics.
Our method is capable of filling large inpainting regions, oftentimes exceeding the quality of comparable methods for high-resolution images.
For reference patch look-up we propose to use the same summary statistics that are used in the inpainting process.
\end{abstract}


\section{Introduction} \label{sec:intro}

\begin{figure}[htb!]
\centering
\subfloat[]{
		\includegraphics[width=0.14\textwidth]{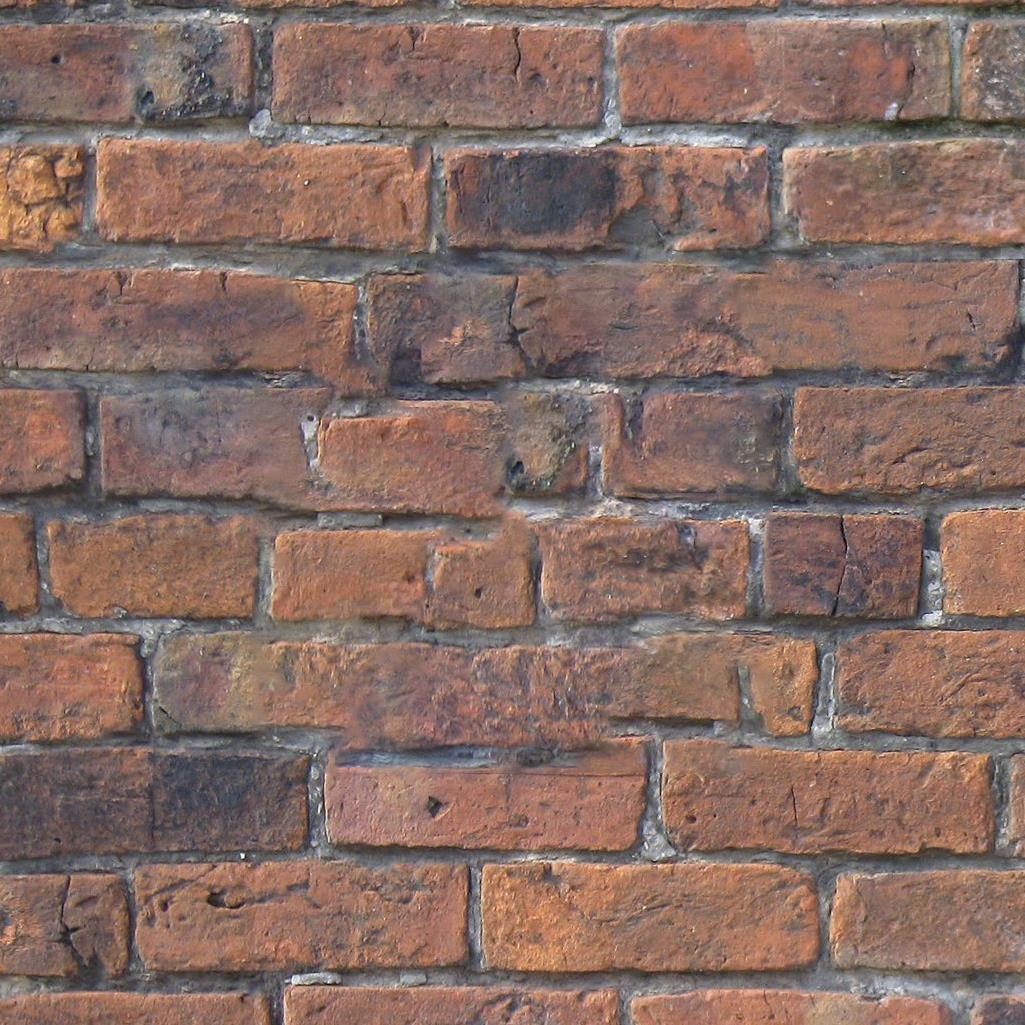}
\label{fig:ps}
}
\subfloat[]{
		\includegraphics[width=0.14\textwidth]{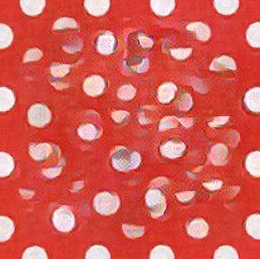}
\label{fig:yang}
}
\subfloat[]{
		\includegraphics[width=0.14\textwidth]{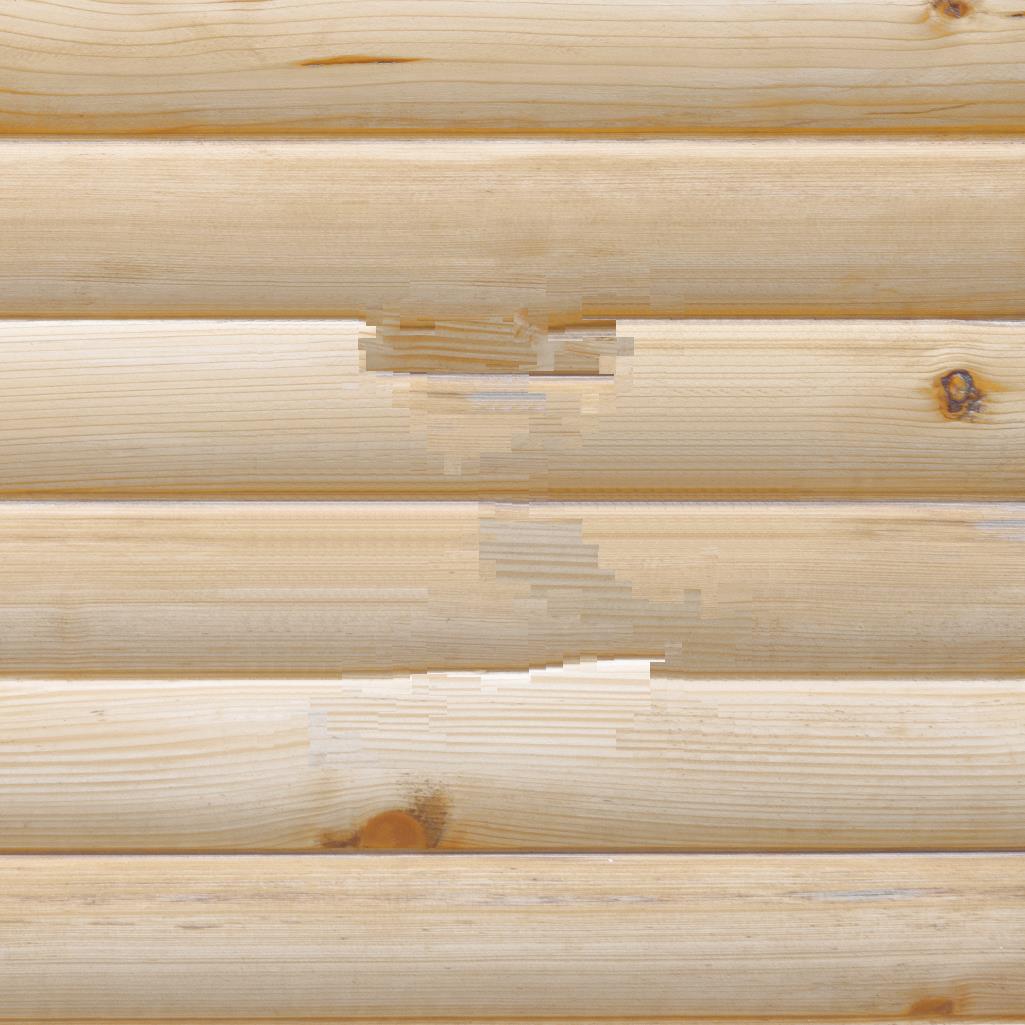}
\label{fig:crimi}
}
\caption{Inpainting results for some of the example textures from Fig. \ref{fig:examples} using the methods of \protect\subref{fig:ps} Photoshop CS7 which is a combination of methods \cite{barnes2009patchmatch} and \cite{wexler2007space}, \protect\subref{fig:yang} the method by Yang et al. \cite{yang2016high}, \protect\subref{fig:crimi} and by Criminisi et al. \cite{criminisi2004region}.} 
\label{fig:compare}
\end{figure}

Image inpainting is the process of filling missing or corrupted regions in images based on surrounding image information so that the result looks visually plausible.
Most image inpainting approaches are based on sampling existing information surrounding the inpainting region, wich is called exemplar-based \cite{criminisi2004region, wexler2007space, kwatra2005texture, efros1999texture, barnes2009patchmatch} inpainting.
Recently machine learning techniques have been applied successfully to the problem of texture synthesis and inpainting \cite{li2016combining, gatys2016image, johnson2016perceptual, dosovitskiy2016generating}.
First introduced by Gatys et al. in \cite{gatys2015texture} texture synthesis CNNs have been shown to surpass well-known methods like the one by Portilla et al. \cite{portilla2000parametric} for many textures.
Wallis et al. \cite{wallis2017parametric} recently showed that artificial images produced from a parametric texture model closely match texture appearance for humans. Especially, the CNN texture model of \cite{gatys2015texture} and the extension by Liu \cite{liu2016texture} are able to capture important aspects of material perception in humans. For many textures the synthesis results are indiscriminable under foveal inspection.
Other methods like the ones by Phatak et al. \cite{pathak2016context} and Yang et al. \cite{yang2016high} train auto-encoder-like networks, called context-encoders, for inpainting.
Inpainting methods using neural networks still suffer two main drawbacks: Due to limited computational resources they are restricted to small inpainting regions and results often lack details and are blurry.
For high-resolution textures the inpainting result not only needs to reproduce texture details but also global structure.
Applying details after a first coarse inpainting step distorts global statistics.
Fig. \ref{fig:compare} shows some examples where well-known inpainting methods fail to reproduce global and local structure. 
\vspace{-1em}
\vspace{-0.5em}

To resolve the outlined issues we propose an inpainting approach that produces results that reproduce global statistics and contain blur-free details.
We fill the inpainting region by synthesis of new texture patch by patch, which enables us to process high-resolution textures.
Our inpainting approach creates a smooth transition between the sampling and the inpainting region as well as between patches.
Out setup is able to shift focus from optimizing detail to global statistics on different levels of resolutions.\\ 
Sections of this paper are arranged as follows.
The process of texture synthesis by CNNs is then explained in Sec. \ref{sec:texsynth}.
In Sec. \ref{sec:methods} we present our inpainting approach, followed by an experimental evaluation in Sec. \ref{sec:results}. We conclude in Sec. \ref{sec:concl}.
\vspace{-1em}
\section{Texture synthesis} \label{sec:texsynth}
First introduced by Gatys et al. \cite{gatys2015texture} CNN texture synthesis uses summary statistics derived from filter responses of convolutional layers, the feature maps, to synthesize new texture.
In a first step some vectorized texture $\x$ of size $P$ is presented to the analysis CNN.
Based on the resulting feature maps one can compute the Gramians which are spatial summary statistics.
The Gramian of some network layer $l$ is defined as
\begin{equation}
\G^l_{ij} = \sum_{k} \F^l_{ik}  \F^l_{jk}, \label{eq:gramian}
\end{equation}
where $\F^l_{ik}$ is feature map $i$ and $\F^l_{jk}$ feature map $j$ at location $k$ given input $\x$.
These inner products of filter activations of different layers are then used to define a synthesis loss
$$
\Lb_s(\x, \hat{\x}) = \sum_{l=0}^L \frac{1}{2N^2_lM^2_l} \sum_{i,j}(\G^l_{ij} - \hat{\G}^l_{ij})^2,
$$
with $N_l$ feature maps of size $M_l$ at layer $l$.
Here $\hat{\G^l}_{ij}$ are the Gramians of a synthesis CNN.
Based on this loss some randomly initialized input vector $\hat{\x}$ of the synthesis CNN is optimized to satisfy statistics derived from the analysis CNN. 
Since Gramians average over feature map positions this leads to a loss of global texture coherence.
Berger and Memisevic \cite{berger2016incorporating} introduce a second cross-correlation loss by computing Gramians between feature maps $\F^l$ and a spatial translation $T$ of the feature maps $T(\F^l)$.
By discarding either rows or columns of feature maps one can now compute correlations of features at some location $k=(x,y)$ and a shifted location $T_{x,+\delta}(k)=(x+\delta,y)$ or $T_{y,+\delta}(k)=(x,y+\delta)$.
The horizontally translated Gramian becomes
\begin{equation}
G^l_{x,\delta,ij} = \sum_{k} T_{x,+\delta}(F^l_{ik}) T_{x,-\delta}(F^l_{jk}), \label{eq:ccgram}
\end{equation}
and $G^l_{y,\delta,ij}$ analogous.
The cross-correlation loss $\Lb_{cc}$ for an arbitrary shift $\delta$ is defined as
$$
\Lb_{cc}(\x, \hat{\x}) = \sum_{l,i,j}\frac{(\G^l_{x,\delta,ij} - \hat{\G}^l_{x,\delta,ij})^2 + (\G^l_{y,\delta,ij} - \hat{\G}^l_{y,\delta,ij})^2}{4N^2_lM^2_l} .
$$
The combined loss is then defined as 
$$
\Lb_{s,cc}(\x, \hat{\x}) = w_s \Lb_s + w_{cc} \Lb_ {cc},
$$
with weight factors $w_s$ and $w_{cc}$.

\section{Patch-based texture synthesis for \\ image inpainting} \label{sec:methods}

\subsection{Patch-based texture synthesis}

\begin{figure*}[htb!]
\centering
		\includegraphics[width=0.9\textwidth]{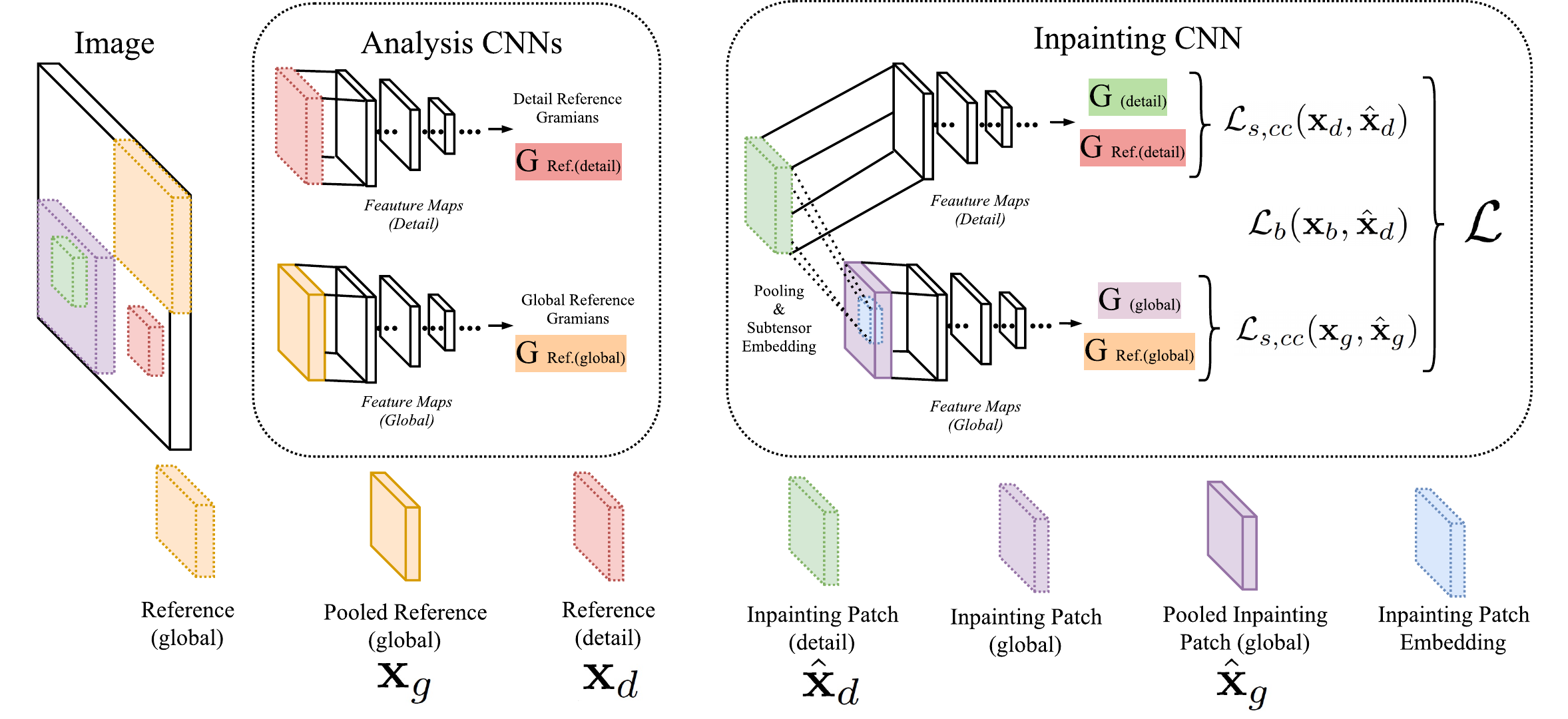}
\caption{Scheme of our proposed inpainting setup. On the top left the inpainting image together with important image regions is shown. Under "Analysis CNN" the generation of detail as well as reference Gramians is shown. On the top right our "Inpainting CNN" together with the resulting loss terms is shown. The inpainting patch $\hat{\x}_d$ is input to the \textit{detail branch} (top) as well as, after embedding, the \textit{global branch} (bottom). A legend of the involved image elements is given at the bottom.}
\label{fig:net}
\vspace{-1em}
\end{figure*}

Given some image with high-resolution, uncorrupted texture $\Phi$ we propose the application 
of the synthesis method introduced in Sec. \ref{sec:texsynth} on different scales of 
resolution to fill the inpainting region $\Omega$ (Fig. \ref{fig:leo3}). A schematic 
overview of our setup is given in Fig. \ref{fig:net}.
We propose to inpaint region $\Omega$ patch by patch with each patch satisfying global as 
well as detail statistics. For this purpose, we define a texture loss function that 
simultaneously evaluates the quality of the synthesized patch $\hat{\x}_d$ in native 
resolution as well as the quality of an embedding of $\hat{\x}_d$ 
into a pooled window of its surroundings $\hat{\x}_g$ capturing global information.
$\hat{\x}_g$ is initialized 
with a $Q$-times average-pooled window of the image so that this window fully contains 
$\Omega$ and the boundary $\Psi$.
$Q$ average-pooling-layers are introduced in-between $\hat{\x}_d$ and $\hat{\x}_g$ so 
that $\hat{\x}_d$ can become a subtensor of $\hat{\x}_g$ at the correct (pooled) position.
Depending on the size of $\Omega$, $Q$ needs to be adjusted as a parameter before inpainting.
Before generating the next patch $\hat{\x}_d$ at a new location we update $\Omega$ with the 
synthesis result in  $\hat{\x}_d$ and reinitialize $\hat{\x}_g$.
Only $\hat{\x}_d$ is optimized in the synthesis process.


For the synthesis as described in Sec. \ref{sec:texsynth}, suitable reference textures 
$\x_d$ and $\x_g$ are needed. We will describe the reference patch look-up in 
Sec.~\ref{sec:patchdist}. While $\x_g$ needs to be initialized only once at the beginning of the inpainting process, $\x_d$ is reinitialized with a new reference for every new position of the inpainting patch $\hat{\x}_d$.
We further define a boundary loss, that limits the optimization of region $\Psi$ inside $\hat{\x}_d$ in the input domain.
We define the boundary loss as
\begin{equation}
\Lb_{b}(\x, \hat{\x}) = \frac{1}{P} \sum (\m \x  - \m \hat{\x})^2, \label{eq:bloss}
\end{equation}
where the binary mask $\m$ equals $0$, if $\m_i \in \Omega$, and $1$ otherwise.
The combined loss over both branches together with boundary loss becomes
$$
\Lb = w_d \Lb_{s,cc}(\x_d, \hat{\x}_d) + w_g \Lb_{s,cc}(\x_g, \hat{\x}_g) + w_b \Lb_b(\x_b, \hat{\x}_d),
$$
where $w_d$, $w_g$, and $w_b$ are weight terms.
$\x_b$ is initialized with $\hat{\x}_d$ before optimization and does change for each new position of $\hat{\x}_d$.  

\subsection{Patch distance by Gramians} \label{sec:patchdist}
For the synthesis of patch $\hat{\x}_d$ ,suitable reference patches $\x_d$, and $\x_g$ are 
needed. The initial $\hat{\x}_d$ is a window of the image containing parts of $\Psi$ as well 
as parts of $\Omega$ while $\hat{\x}_g$ completely contains $\Psi$ and $\Omega$.
One now has to find closest patches from $\Phi$ matching $\Psi$ inside $\hat{\x}_d$, and 
$\hat{\x}_g$ as candidates for $\x_d$, and $\x_g$.
Instead of the MSE, we propose to use the distance of texture Gramians as a similarity measure.
Since values inside $\Omega$ are unknown we propose masking $\Omega$ for each individual 
feature map to remove $\Omega$-related correlations from any of the resulting Gramians.
Because the network input up to some layer $l$ has passed through both pooling and 
convolutional layers we need to adapt the feature map masks to compensate for these operations.
In a first step, the initial binary mask $\m$ from Eq. (\ref{eq:bloss}) needs to be adapted 
in size to account for the pooling layers.
This is done by applying each pooling step of the CNN that has been applied up to layer 
$l$ to the mask $\m^l$ which is responsible for masking feature maps $\F^l$.
In a second step, one needs to account for the propagation of $\Omega$ into $\Psi$ by 
convolutions.
Masks $\m^l$ also need to account for propagation of $\Omega$ into $\Psi$ due to convolution.
Simply discarding the affected values by setting them to zero in $\m^l$ for each 
convolutional layer is too restrictive and would lead to masks with all values zero in 
later layers.
We propose to expand $\Omega$ by a smaller individual number of pixels $\e^l$ for each 
convolutional layer (see Sec. \ref{sec:results}).
In our experiments this expansion has proven to be sufficient for compensation.\\
Taking these considerations into account we define our patch distance as
$$
\Delta_\G(\x,\hat{\x}) = \sum_{l,i,j}(\sum_{k} \m^l\F^l_{ik}\F^l_{jk} - \sum_{k} \m^l\hat{\F}^l_{ik}\hat{\F}^l_{jk})^2.
$$

\subsection{Inpainting}

\begin{figure}[htb!]
\centering
\subfloat[]{
		\includegraphics[width=0.14\textwidth]{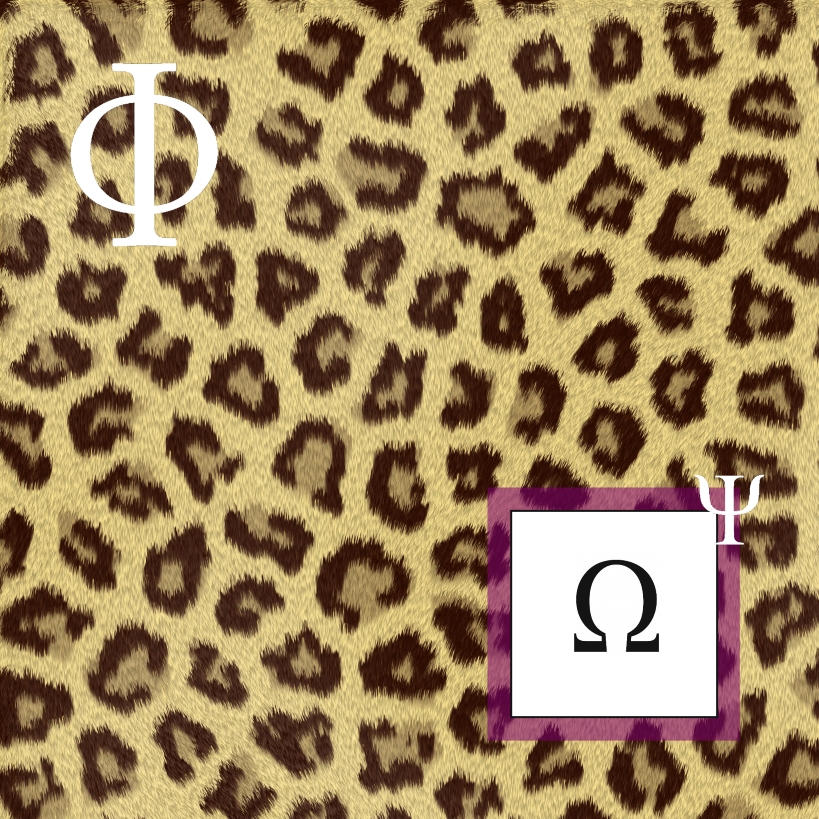}
\label{fig:leo3}
}
\subfloat[]{
		\includegraphics[width=0.14\textwidth]{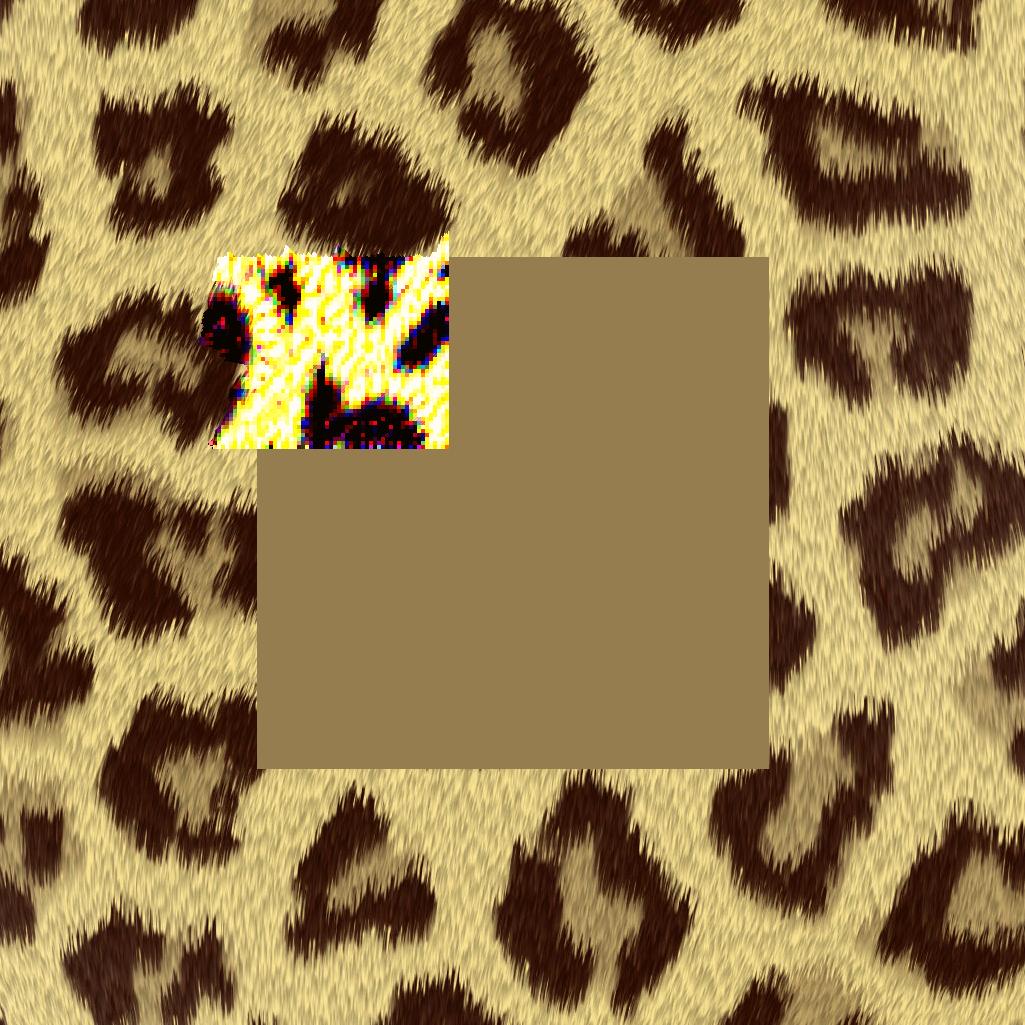}
\label{fig:leo4}
}
\subfloat[]{
		\includegraphics[width=0.14\textwidth]{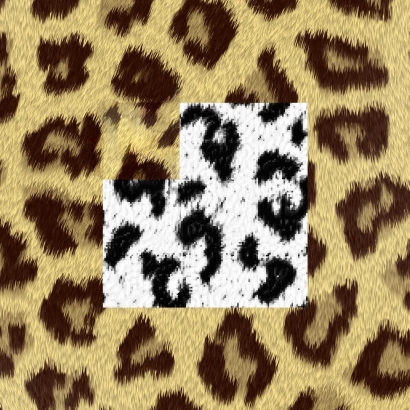}
\label{fig:leo5}
}
\caption{\protect\subref{fig:leo3} Example image (2048x2048px) with inpainting region $\Omega$, boundary $\Psi$ and texture $\Phi$.  
\protect\subref{fig:leo4} First patch of the coarse inpainting step. 
\protect\subref{fig:leo5} Fine inpainting of texture detail after coarse inpainting.}
\label{fig:leo}
\end{figure}

For inpainting we propose a coarse to fine inpainting process with two steps.
At each stage $\hat{\x}_d$ is optimized by applying L-BFGS-B \cite{zhu1997algorithm}.
We initialize each color channel in region $\Omega$ with the corresponding color channel mean from $\Phi$.
In the coarse inpainting step we focus on optimizing global statistics by setting $w_d = 0$, $w_g = 1$.
This leads to $\hat{\x}_d$ satisfying global statistics but at low resolution.
Pooling larger input regions introduces color artifacts since loss is shared among pooled pixels as can be seen in Fig. \ref{fig:leo4}.
We eliminate these color artifacts by converting $\Omega$ to greyscale (see Fig. \ref{fig:leo5}) with RGB weights $r=0.212$, $g=0.7154$, $b=0.0721$.
Only this structure is used for initialization of the second stage.
In the fine inpainting step we set $w_d=1$ and $w_g$ to a value in the range of $[0.01, 0.1]$.
This ensures focus on the optimization for detail statistics through the \textit{detail branch} while constraining the optimization to also maintain global texture statistics.\\
For our approach inpainting order is not important as long as the first patch overlaps with $\Psi$ and consecutive patches overlap.
Overlapping a patch by $\frac{1}{4}$ of its own size with surrounding texture has proven to be sufficient for smooth boundary transition.
We chose to fill $\Omega$ in a top to bottom, left to right fashion.
To ensure a smooth transition in-between patches we apply image quilting \cite{efros2001image} on overlaps.
As a result of our experiments we set $w_s=1e6$ and $w_{cc}=1e7$ for inpainting of 8-bit color images.
Choosing $w_b$ in the range $[5,25]$ has shown to be sufficient.
The large difference between Gramian-based loss weights and weights related to loss in pixel space results from different value ranges.
\vspace{-0.5em}
\vspace{-2em}

\section{Experimental Evaluation} \label{sec:results}
\begin{figure}[htb!]
\centering
\subfloat[]{
		\includegraphics[width=0.11\textwidth]{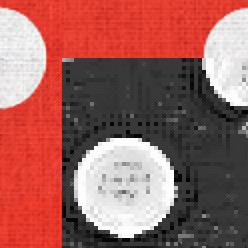}
\label{fig:polka7}
}
\subfloat[]{
		\includegraphics[width=0.11\textwidth]{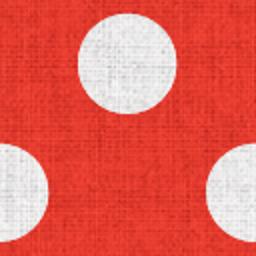}
\label{fig:polka8}
}
\subfloat[]{
		\includegraphics[width=0.11\textwidth]{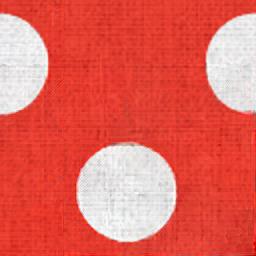}
\label{fig:polka9}
}
\caption{\protect\subref{fig:polka7} Inpainting patch $\hat{\x}_d$. \protect\subref{fig:polka8} Closest reference patch from $\Phi$. \protect\subref{fig:polka9} Inpainting result.}
\label{fig:references}
\end{figure}

\begin{figure}[htb!]
\centering
\hspace{0.1em}
		\includegraphics[width=0.15\textwidth]{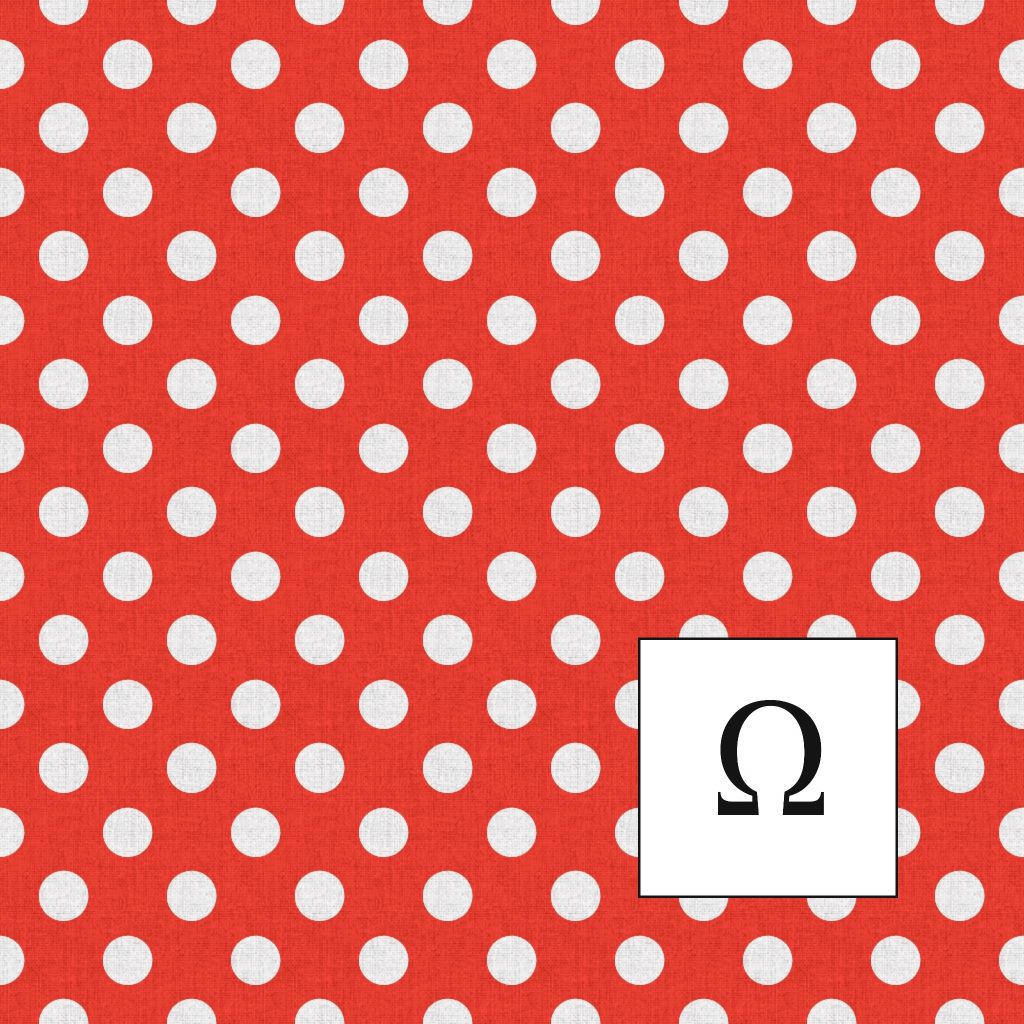}
		\includegraphics[width=0.15\textwidth]{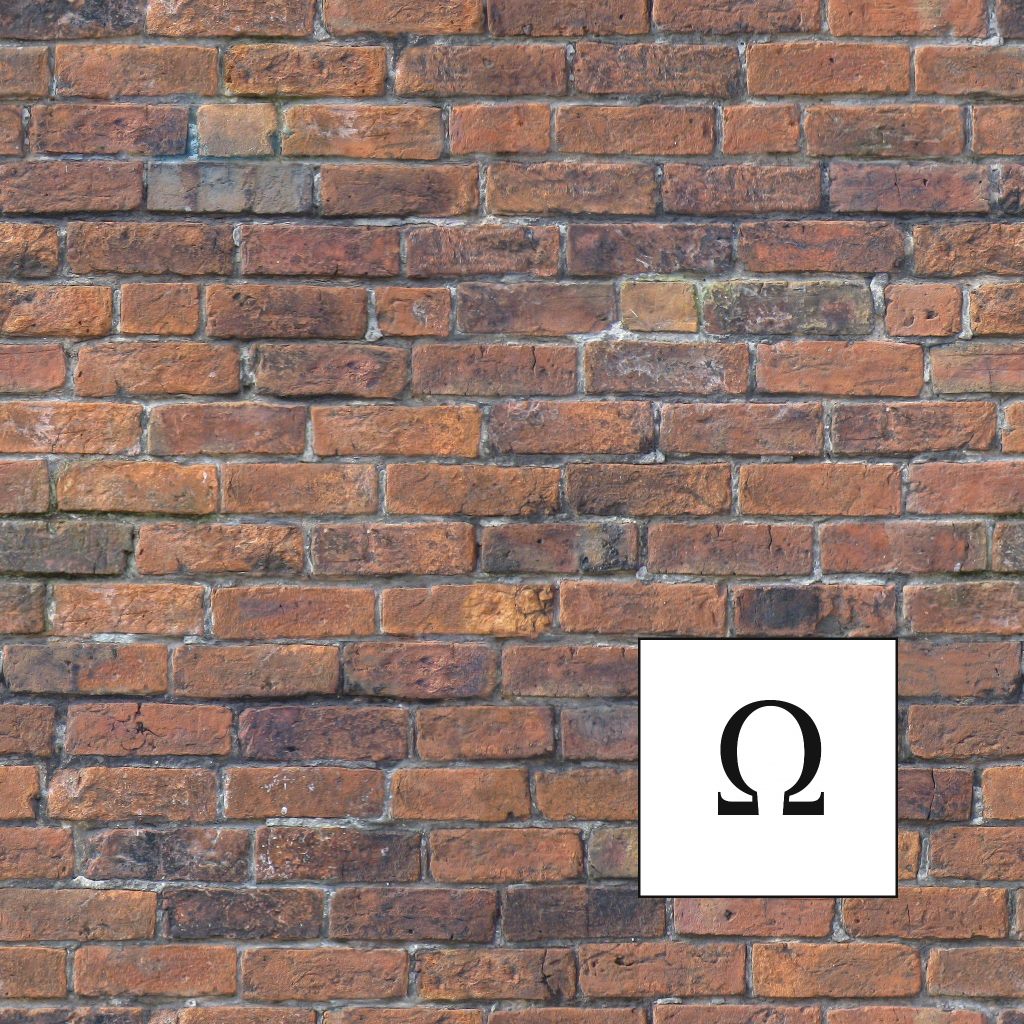}
		\includegraphics[width=0.15\textwidth]{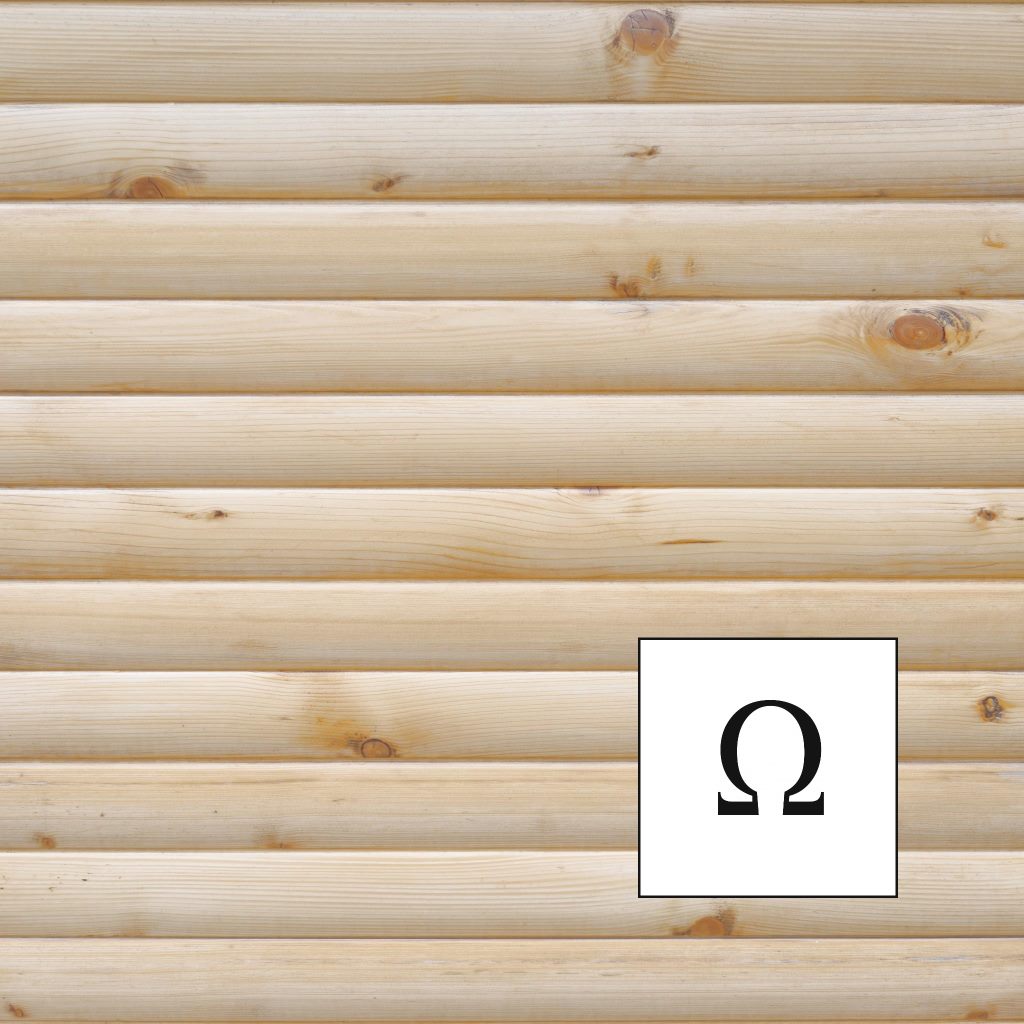}
\vspace{0.5em}
\newline
		\includegraphics[width=0.15\textwidth]{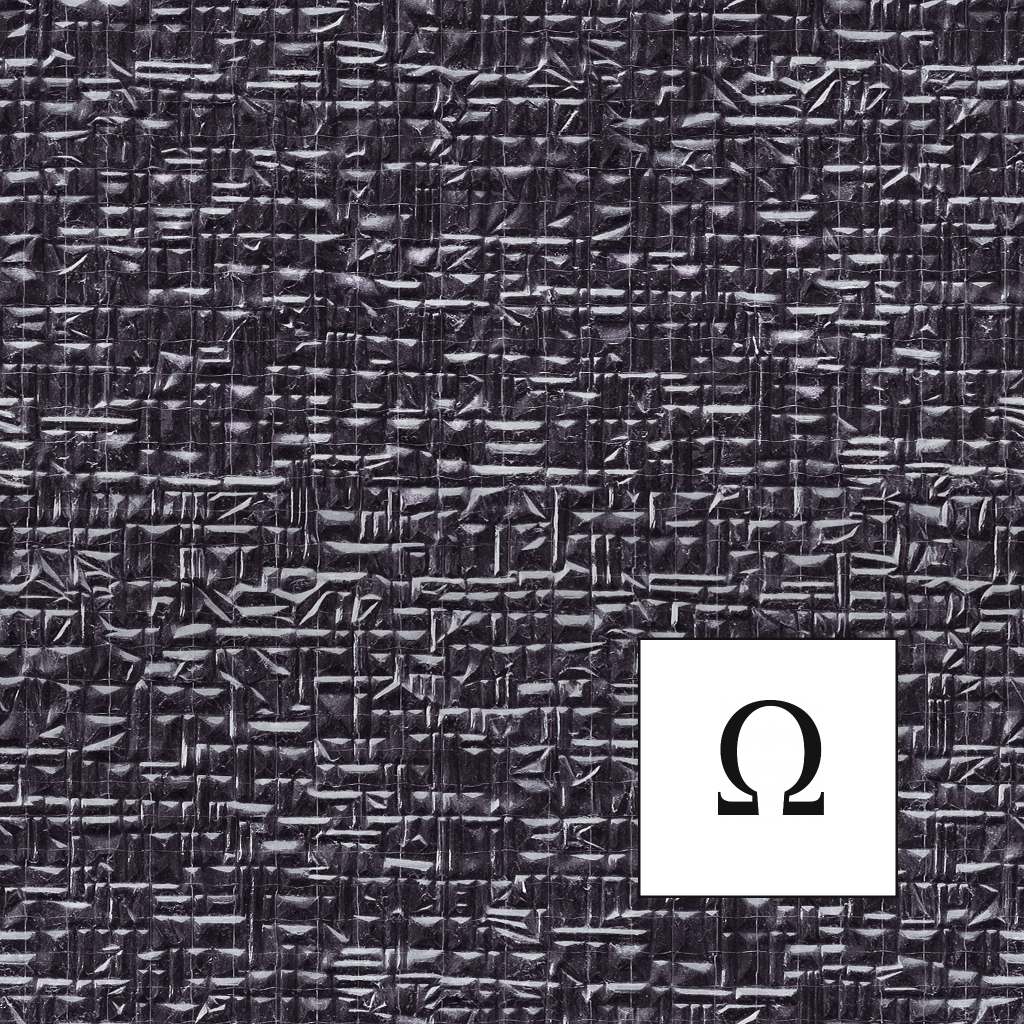}
		\includegraphics[width=0.15\textwidth]{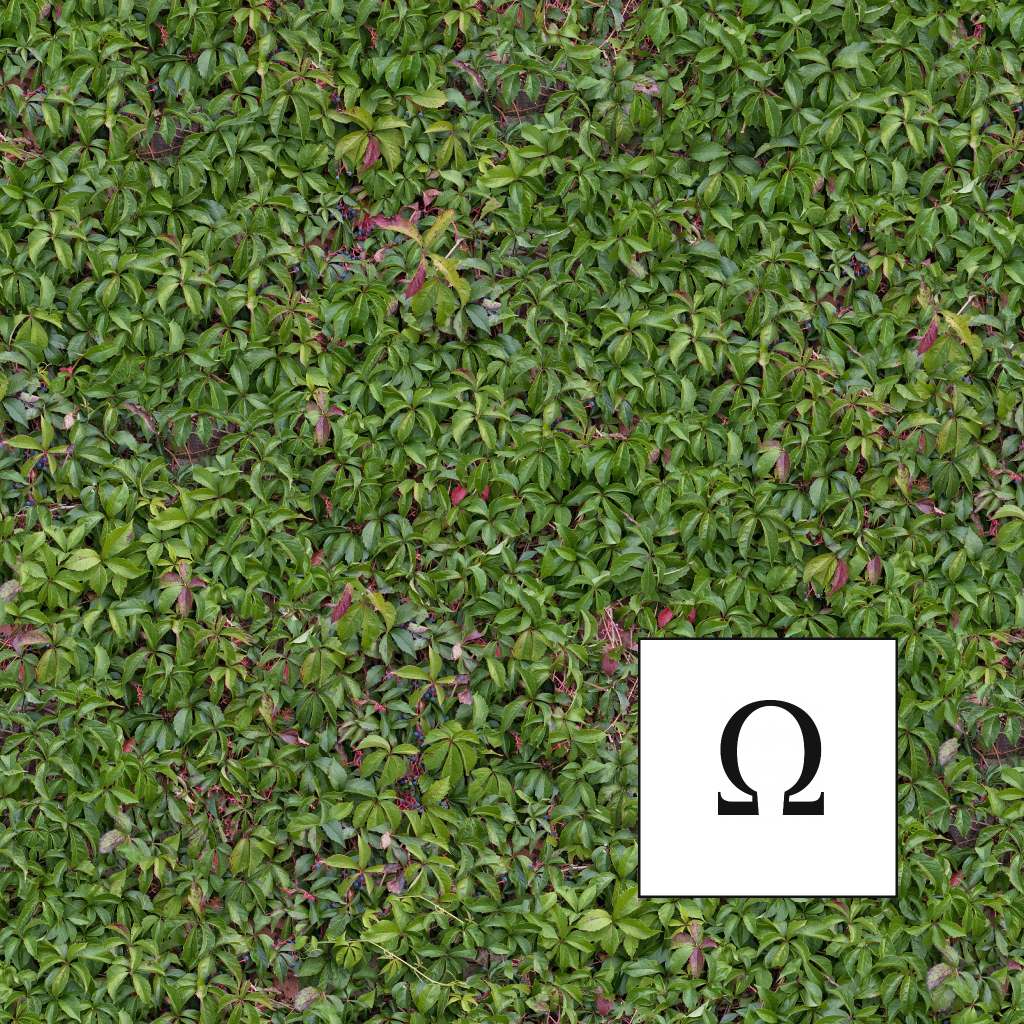}
		\includegraphics[width=0.15\textwidth]{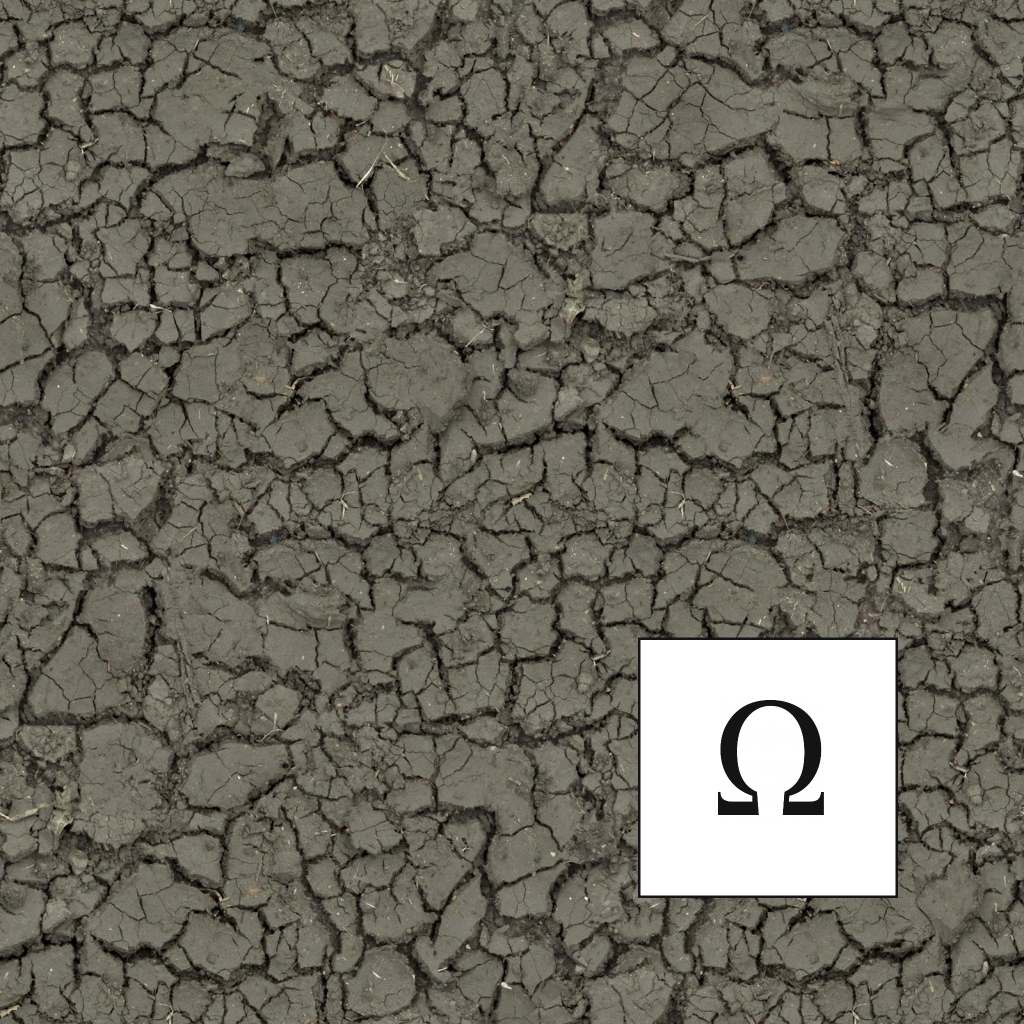}
\caption{Examples for evaluation with inpainting region $\Omega$.}
\label{fig:examples}
\end{figure}

\begin{figure}[htb!]
\centering
		\includegraphics[width=0.23\textwidth]{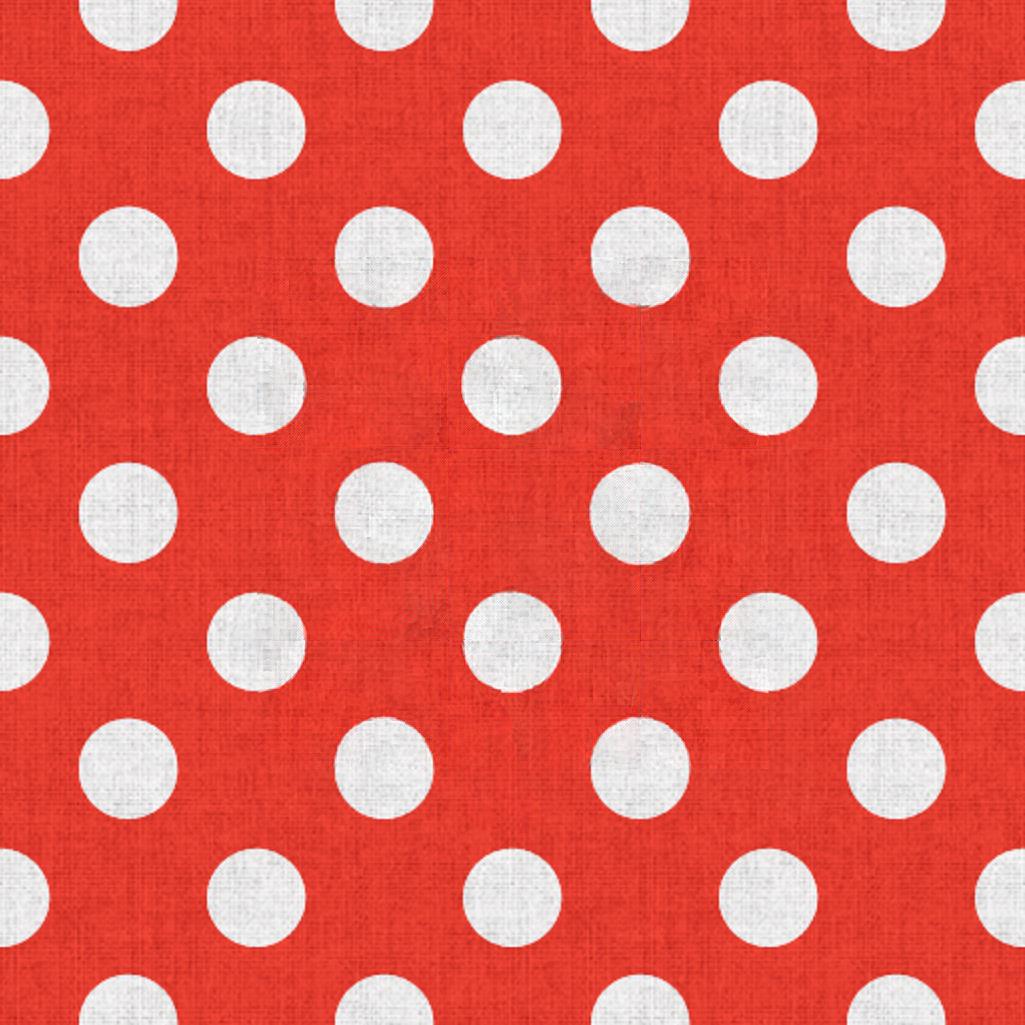}
		\hspace{0.05em}
		\includegraphics[width=0.23\textwidth]{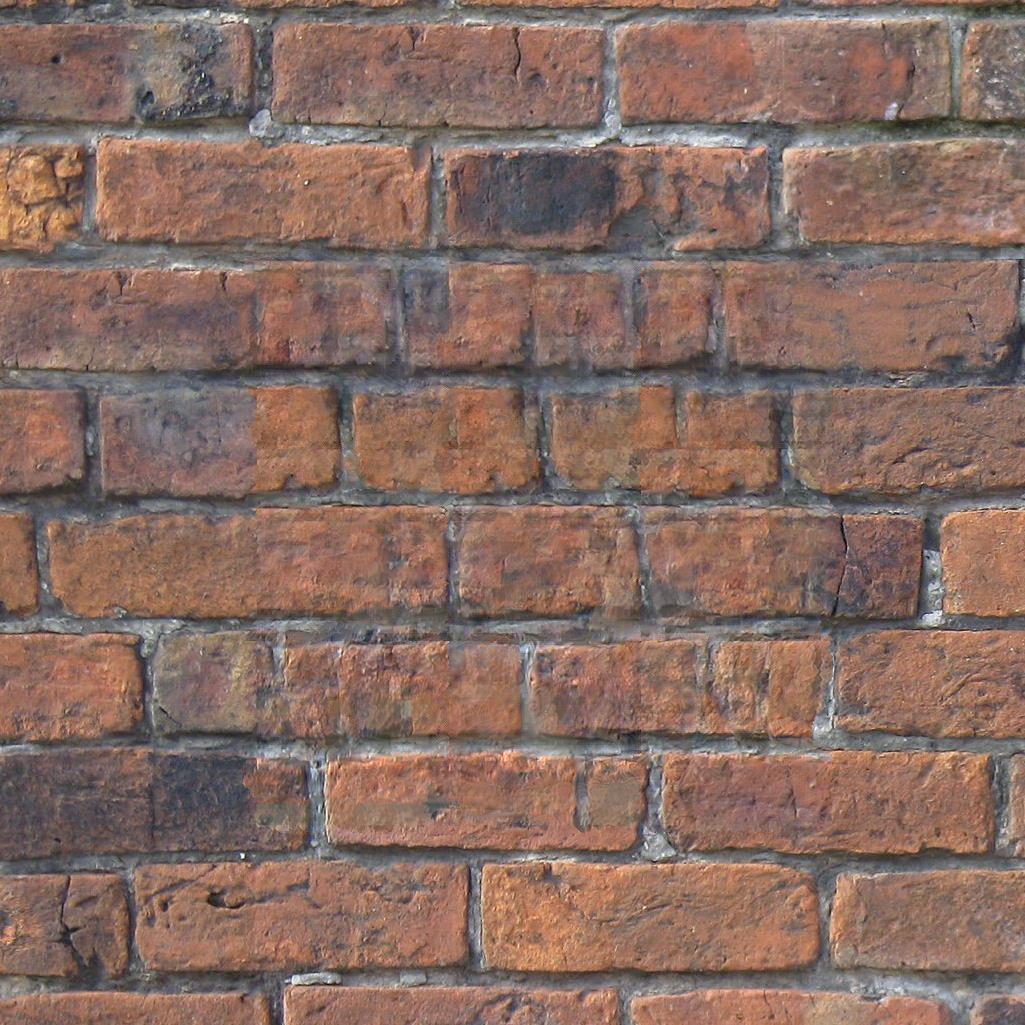}
		\vspace{0.5em}
\newline
		\includegraphics[width=0.23\textwidth]{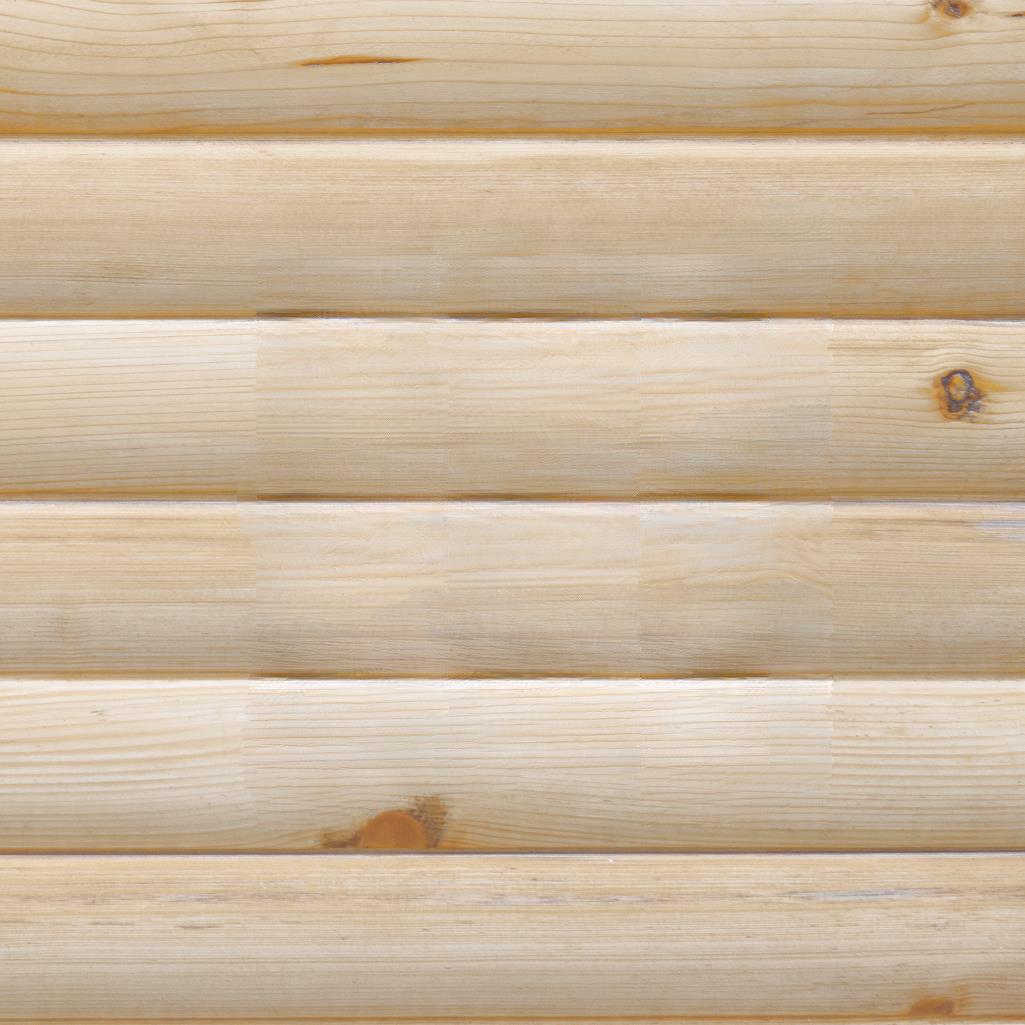}
		\hspace{0.05em}
		\includegraphics[width=0.23\textwidth]{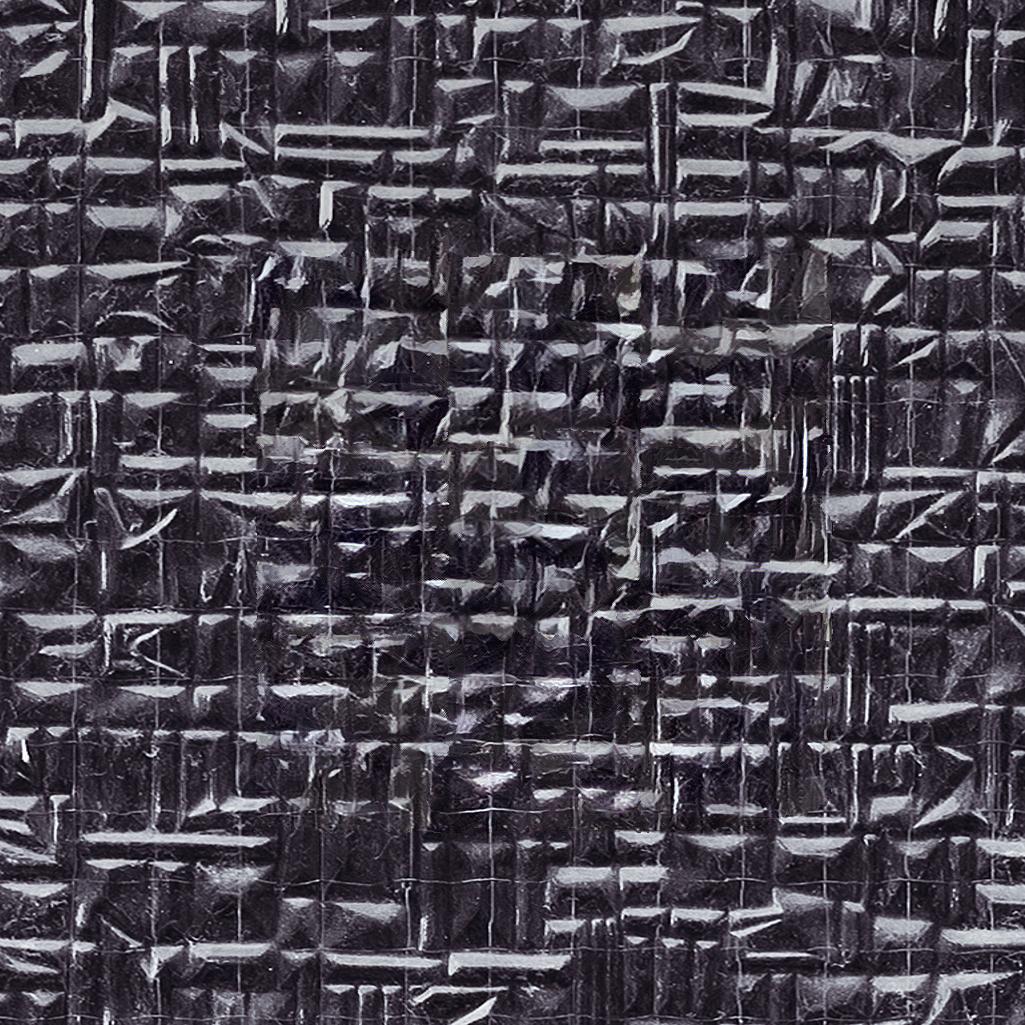}
		\vspace{0.5em}
\newline
		\includegraphics[width=0.23\textwidth]{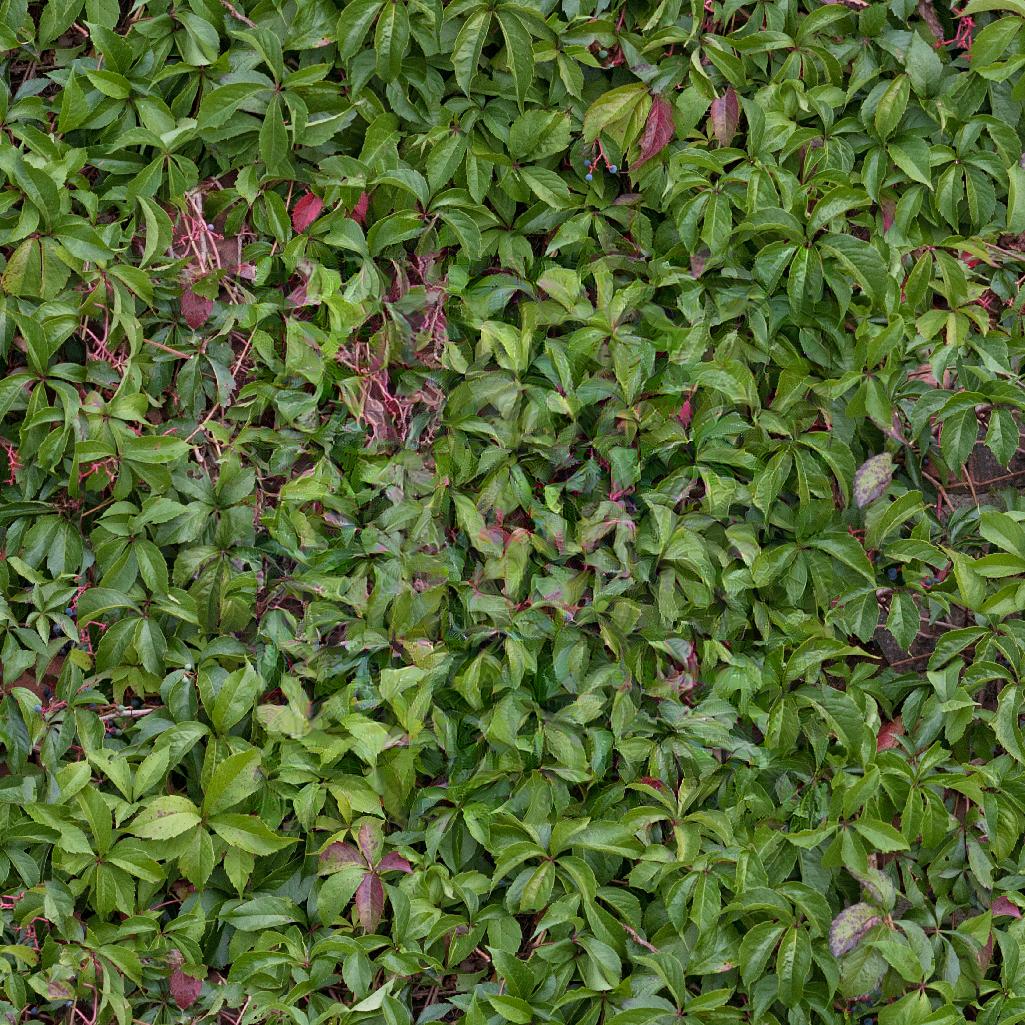}
		\hspace{0.05em}
		\includegraphics[width=0.23\textwidth]{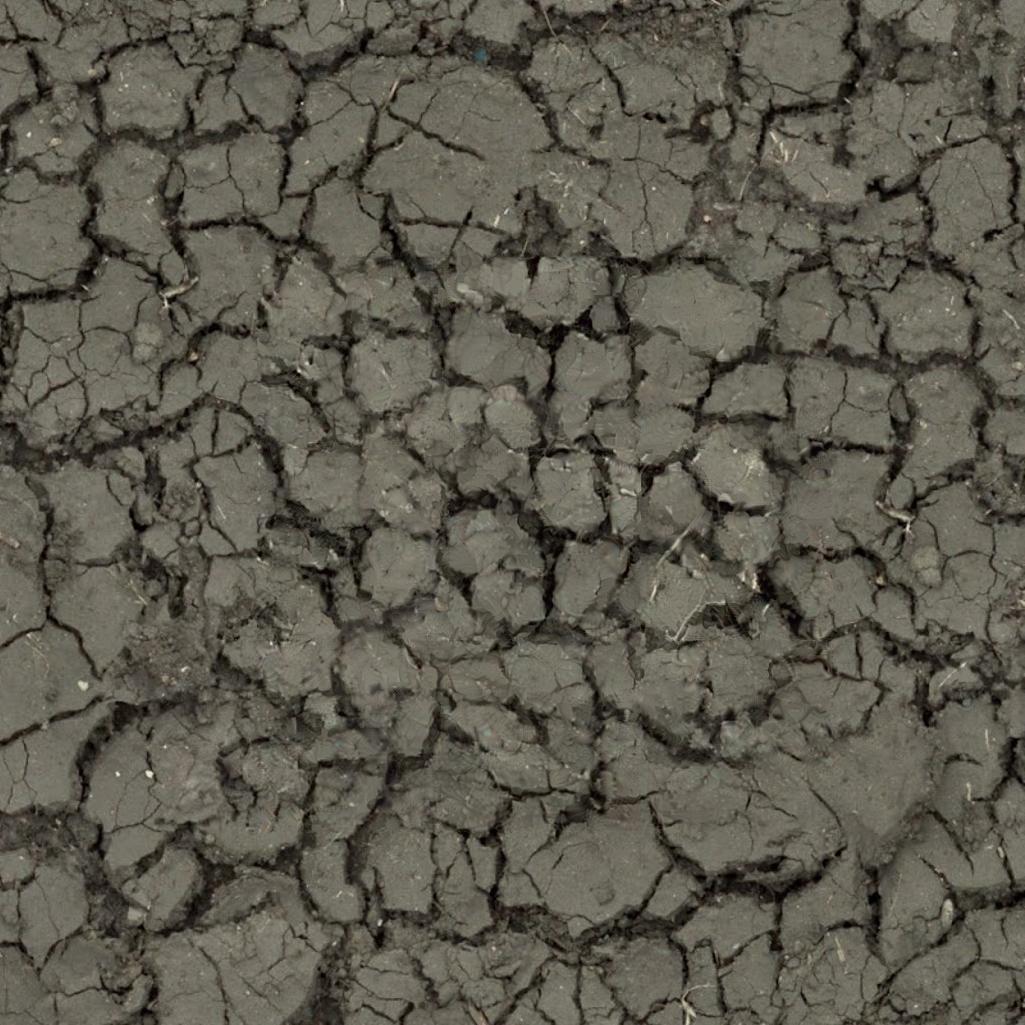}
\caption{Closeup on results using our method to inpaint region $\Omega$ from Fig. \ref{fig:examples}.}
\label{fig:results}
\vspace{-1.4em}
\end{figure}

We present inpainting results of exemplar high resolution textures.
All textures have a resolution of $2048$x$2048$px while the inpainting region $\Omega$ is of size $512$x$512$px.
We use ImageNet pre-trained VGG-19 CNNs for analysis as well as synthesis with input size $256$x$256$px.
We use layers \textit{conv1\_1}, \textit{pool1}, \textit{pool2}, \textit{pool3} and \textit{pool4} for computing global as well as detail statistics.
For very stochastic textures we propose to use \textit{pool3}, \textit{pool4} and \textit{pool5} to compute global statistics since this leads to improved texture scale in the coarse inpainting step.
For patch distance computation we define pixel expansions $\e = (1, 1, 2, 3, 2)$, and for shift $\delta$ of translated Gramians $\G_{x,\delta}^l$ and $\G_{y,\delta}^l$ we define $\boldsymbol{\delta} = (6, 6, 5, 4, 3)$. We use $Q=2$ pooling layers.
To find suitable reference patches $\x_d$ and $\x_g$ region $\Phi$ is searched at a step size of $64$px.
Inpainting of the exemplar textures was done using a Nvidia GeForce 1080 Ti and took roughly $8$ min strongly depending on the number of iterations of the L-BFGS-B optimization.
In Fig. \ref{fig:results} we present results of our inpainting approach for inpainting $\Omega$ of the example textures in Fig. \ref{fig:examples}.
While many methods have difficulties maintaining global as well as local texture characteristics our results look reasonable on both scales.
Using the difference of masked Gramians as a metric for patch distance has major benefits for our inpainting approach over using simple MSE.
Since we are not dependent on reference textures $\x_d$ or $\x_g$ exactly matching $\Psi$ inside $\hat{\x}_d$ or $\hat{\x}_g$ in terms of MSE, we can reduce the number of samples taken from $\Phi$ in reference patch look-up.
Due to the averaging of feature information inside Gramians, global spacial information is lost.
This enables the Gramian to represent texture invariant to rotation and translation to some degree (see Fig. \ref{fig:references}).
Because our loss term $\Lb$ is based on the difference of Gramians this further ensures that $\Psi$ inside $\hat{\x}$ already satisfies target statistics to some extent.
\vspace{-0.5em}
\vspace{-0.5em}
When choosing $w_d$ and $w_g$ one needs to be aware of the trade-off introduced.
While higher $w_g$ ensures persistence of global statistics it also introduces artifacts as a result of pooling $\hat{\x}_d$ before subtensor embedding and vice versa. 
Higher $w_d$ lays larger emphasis on details while possibly violating global structure.
This trade-off is further influenced by the number of poolings $Q$.
\vspace{-0.5em}
\section{Conclusion} \label{sec:concl}
In this work, we presented a new CNN-based method for inpainting that can be applied to large-scale, high-resolution textures. Texture analysis and inpainting are done on two scales, one for global structure and one for details. This avoids the problems of blurry or missing details from which previous CNN approaches suffered while plausibly continuing global image structure. In principle, our network architecture can be extended to include a hierarchy of more than two interacting scales. The design of such a multi-resolution architecture could be an interesting line of research that we plan to pursue in the future.

\bibliographystyle{IEEEbib}
\bibliography{mybibfile}

\end{document}